\documentclass{article}

\usepackage[preprint]{neurips_2026}
\makeatletter
\renewcommand{\@notice}{}
\makeatother

\usepackage[utf8]{inputenc}
\usepackage[T1]{fontenc}
\usepackage{hyperref}
\usepackage{url}
\usepackage{booktabs}
\usepackage{amsfonts}
\usepackage{amssymb}
\usepackage{nicefrac}
\usepackage{microtype}
\usepackage{xcolor}

\hypersetup{
  colorlinks,
  linkcolor={red!50!black},
  citecolor={blue!50!black},
  urlcolor={blue!80!black}
}

\usepackage{mathtools}
\usepackage{amsthm}

\usepackage{graphicx}
\usepackage{caption}
\usepackage{subcaption}
\usepackage{enumitem}
\usepackage{algorithm}
\usepackage{algpseudocode}
\usepackage{wrapfig}
\usepackage{needspace}
\usepackage[capitalize]{cleveref}

\theoremstyle{plain}
\newtheorem{theorem}{Theorem}

\newtheorem{proposition}[theorem]{Proposition}

\theoremstyle{definition}
\newtheorem{assumption}{Assumption}
\newtheorem{definition}{Definition}
\theoremstyle{remark}

\crefname{assumption}{Assumption}{Assumptions}
\Crefname{assumption}{Assumption}{Assumptions}
\def\spearman{{\rho}^{\scriptstyle\texttt{\upshape{s}}}}
\def\pearson{{\rho}^{\scriptstyle\texttt{\upshape{p}}}}
\def\empsp{\hat{\rho}^{\scriptstyle\texttt{\upshape{s}}}}
\def\emppe{\hat{\rho}^{\scriptstyle\texttt{\upshape{p}}}}
\def\unbpe{\hat{\rho}^{\scriptstyle\texttt{\upshape{p,corr}}}}
\def\unbsp{{\hat{\rho}^{\scriptstyle\texttt{\upshape {s,unb}}}}}
\def\E#1{\mathbb{E}\left[{#1}\right]}
\def\Parg#1{\mathbb{P}\left[{#1}\right]}
\def\Var{\operatorname{Var}}
\def\Cov{\operatorname{Cov}}

\def\sign{\operatorname{sign}}
\def\rank{\operatorname{rank}}
\newcommand{\bE}{\mathbb{E}}
\newcommand{\Corr}{\mathrm{Corr}}
\newcommand{\bigO}{\mathcal{O}}

\newcommand{\piS}{\pi_{\texttt{S}}}
\newcommand{\hatpiS}{\hat{\pi}_{\texttt{S}}}

\title{Efficient and Accurate Causal Validation of\\ Training Data Attribution}

\author{Simon Vary\textsuperscript{1}\thanks{Equal contribution. Work done while AMH was a MATS Fellow.}
  \quad
  Alireza Mousavi-Hosseini\textsuperscript{2,3,4}\footnotemark[1]
  \quad
  Roger Grosse\textsuperscript{2,3} \\[0.5ex]
  \textsuperscript{1}MATS Research
  \qquad
  \textsuperscript{2}University of Toronto
  \qquad
  \textsuperscript{3}Vector Institute
  \qquad
  \textsuperscript{4}Numurho
}

\begin{document}

\maketitle

\begin{abstract}
    Causal validation of training data attribution (TDA) methods is costly, since evaluation involves retraining a model on many different subsets of the original training set. For example, the Linear Datamodeling Score (LDS), which has become a standard metric for evaluating TDA methods, measures the correlation between TDA predictions and expected retrained model outputs. In this paper, we reduce the amount of computation needed to estimate the LDS along three axes: the number of sampled training subsets, the number of retraining seeds per subset, and the number of models in the attribution ensemble. Each of these factors induces a form of bias in the LDS estimation, and removing these sources of bias significantly reduces the number of retrainings needed to match the same accuracy in LDS estimation. On ResNet-9/CIFAR-10, finite-subset debiasing achieves $3.2\%$ relative RMSE with as few as $4$ subsets, and $32$ subsets match the accuracy of ordinary Spearman using $96$ subsets, measured against a 300-subset reference. On ResNet-9/CIFAR-10, BERT/QNLI and GPT-2/WikiText-2, estimating the training-noise variance allows us to reproduce the standard five-seed LDS at the same accuracy with $2$--$5\times$ fewer retrainings.
\end{abstract}

\section{Introduction}\label{sec:intro}

Training Data Attribution (TDA) methods \citep{Koh2017Influence, Pruthi2020TracIn, Grosse2023EKFAC, Park2023TRAK, Bae2024SOURCE} aim to approximate the causal effect of training data on model behaviour after training. A widely used metric for evaluating the accuracy of their causal predictions is the Linear Datamodeling Score (LDS), which was first introduced by \citet[Definition~2.3]{Park2023TRAK} in the context of TDA and builds on the datamodel framework of \citet{Ilyas2022Datamodels}. LDS measures the correlation, across training subsets, between TDA predictions and the expected outputs of models retrained on those subsets, where the expectation is over training randomness. As such, computing LDS is computationally expensive, since each sample used to estimate the correlation requires multiple full model retrainings, one for every random training seed. This limits evaluation to small scales. LDS has nevertheless become a standard evaluation practice for TDA, and has been adapted even beyond the supervised learning setting in which it was first introduced \citep{Georgiev2023Journey, Bae2024SOURCE, Deng2024Dattri, Quirke2026Bergson}.

\begin{figure}[t]
  \centering
  \includegraphics[width=1\textwidth]{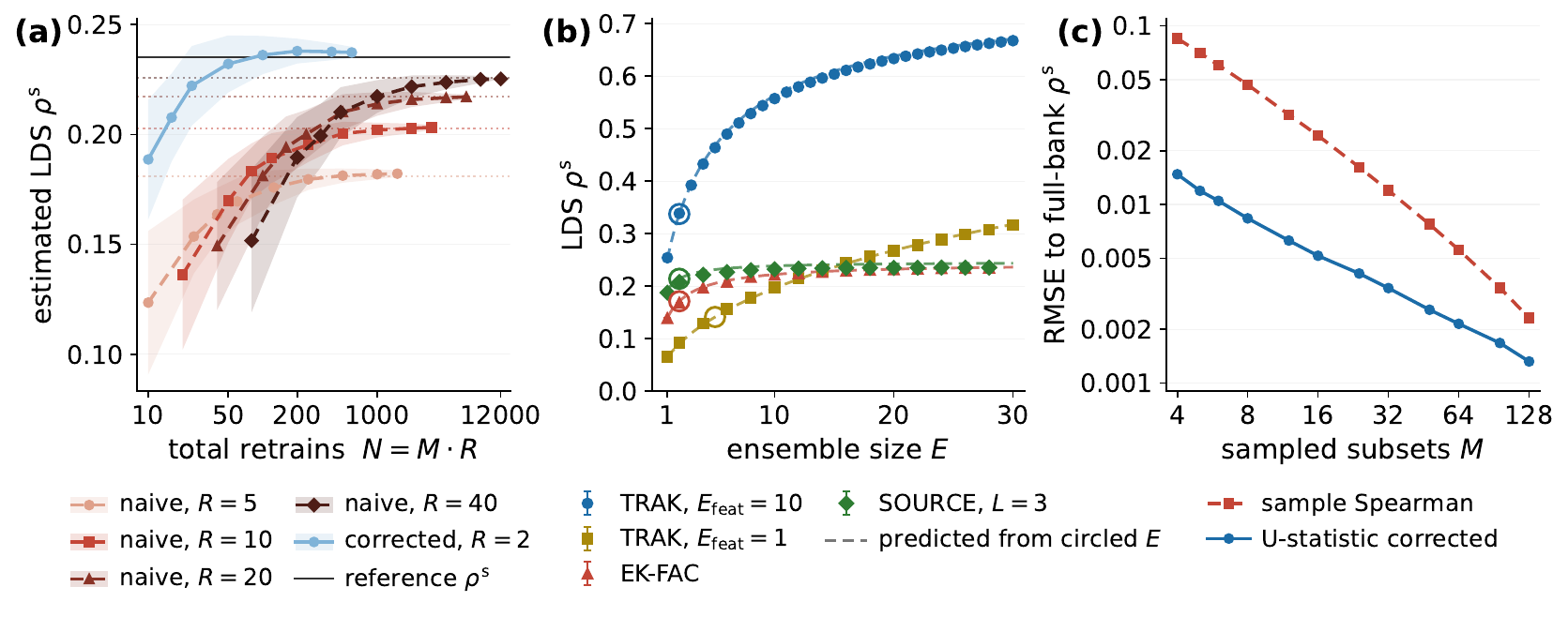}
  \begin{subfigure}{0pt}\phantomsubcaption\label{fig:fig1a}\end{subfigure}\begin{subfigure}{0pt}\phantomsubcaption\label{fig:fig1b}\end{subfigure}\begin{subfigure}{0pt}\phantomsubcaption\label{fig:fig1c}\end{subfigure}\caption{\textbf{Three ways to reduce LDS evaluation cost on CIFAR-10/ResNet-9.} \textbf{(a)} Correcting for training variance approaches the Spearman reference computed from held-out retraining seeds with fewer retrainings, while naive seed averaging plateaus (EK-FAC). Curves show mean estimates $\pm1$ SD over random draws, see \cref{app:experimental-details} for details. \textbf{(b)} The circled small attribution ensembles predict the LDS of larger ensembles (dashed lines). \textbf{(c)} Correcting finite-subset Spearman bias reduces RMSE against the 300-subset reference (TRAK).}
  \label{fig:intro}
\end{figure}

In order to reduce the computational cost of evaluating LDS, we analyze the key sources of error in the LDS estimates and give recipes for reducing each error source at a given level of computation.
The first source of error is randomness in the training procedure (e.g.~from parameter initialization and batch ordering), which results in noise in the ground truth targets themselves. One typically mitigates this noise by averaging over multiple random seeds, but this is expensive as it must be done for \emph{each} data subset. To cheaply reduce this error source, we use multiple random seeds to estimate the retraining-noise variance, and then correct for it when estimating LDS. Empirically, this allows us to estimate the noise-free LDS values reasonably accurately with as few as 2 random seeds per subset (\cref{fig:fig1a}).

A second source of error is the randomness in the TDA method (e.g., due to the choice of checkpoint at which to evaluate the TDA criterion). It has been observed that ensembling TDA methods reduces this randomness and improves their performance \citep{Deng2024Ensembles, Bae2024SOURCE, Park2023TRAK}, but evaluating ensembles can be expensive. By estimating the attribution noise-to-signal ratio and its effect under averaging, we can predict the LDS of large TDA ensembles from as few as two members (\cref{fig:fig1b}).

Finally, a third source of error in LDS comes from estimating the Spearman correlation between TDA predictions and expected retrained model outputs using a finite number of training subsets. The usual sample Spearman estimator has a finite-subset bias, which accounts for a substantial part of the LDS estimation error in our experiments. Correcting for this bias reduces the retraining budget needed for accurate evaluation (\cref{fig:fig1c}).

As a side observation which our analysis depends on but which may be of independent interest, we also observe that the paired TDA predictions and the retrained model outputs are well approximated by a jointly Gaussian distribution. This allows us to convert the corrected Pearson correlation to Spearman correlation, the latter being the standard correlation measure used to compute LDS \citep[Definition 2.1]{Park2023TRAK}.

Our code and precomputed retraining banks are publicly available.\footnote{Code: \url{https://github.com/simonvary/corrected-lds}.\\ \mbox{}\hspace{1.5em} Retraining banks: \url{https://huggingface.co/datasets/corrected-lds/data}.}

\section{Setup: Linear Datamodeling Score}\label{sec:setup}
The Linear Datamodeling Score (LDS) \citep[Definition 2.3]{Park2023TRAK} measures the correlation, across training subsets $S\subset D$, between TDA predictions $G_q(S)$ and expected retrained model outputs $F_q(S)$. Here $q$ denotes a query test point, and the expectation is over training randomness.

We sample $S_1,\ldots,S_M$ independently from
$\piS \coloneqq \mathrm{Unif}\{S\subset D: |S|=\alpha|D|\}$,
and let $\hatpiS$ denote their empirical distribution,
which assigns weight $1/M$ to each sampled subset.
\begin{definition}[Linear Datamodeling Score]\label{def:lds-query}
The empirical LDS is the Spearman correlation between the TDA
predictions and expected retrained model outputs, averaged over
the $Q$ queries:
\begin{equation}\label{eq:lds_sample}
    \empsp_M \coloneqq \frac{1}{Q}\sum_{q=1}^Q \empsp_M(q),
    \qquad
    \empsp_M(q) \coloneqq
    \Corr_{S\sim\hatpiS}
    \left(\rank_M(G_q(S)),\rank_M(F_q(S))\right),
\end{equation}
where $\rank_M$ denotes ranks among the $M$ sampled subsets.

The corresponding population LDS is
\begin{equation}
    \spearman \coloneqq \frac{1}{Q}\sum_{q=1}^Q \spearman(q),
    \qquad
    \spearman(q) \coloneqq
    \Corr_{S\sim\piS}
    \left(H^G_q(G_q(S)),H^F_q(F_q(S))\right),
\end{equation}
where $H_q^G$ and $H_q^F$ are the cumulative distribution
functions of $G_q(S)$ and $F_q(S)$, respectively.
\end{definition}

While the standard LDS uses Spearman correlation \citep{Park2023TRAK}, we derive the noise corrections using Pearson correlation, whose covariance form makes the contribution of the training noise explicit, and later we convert the estimates to Spearman LDS under a joint Gaussian assumption. We define Pearson LDS as
\begin{equation}
    \pearson \coloneqq \frac{1}{Q}\sum_{q=1}^Q \pearson(q), \quad \pearson(q) \coloneqq \Corr_{S\sim\piS}(G_q(S), F_q(S)),
\end{equation}
and the empirical counterpart $\emppe_M$ is defined by replacing $\piS$ with $\hatpiS$. 

\subsection{Statistical Error in LDS Estimation}\label{subsec:challenges}
\paragraph{Stochastic Training Noise.} 
Retraining on the same subset can produce different model outputs because of training randomness. Instead of observing the expected output $F_q(S)$, we observe $f_q(S,\xi)$:
\begin{equation}
    f_q(S, \xi) = F_q(S) + \varepsilon_q(S, \xi),\qquad \text{with}\quad F_q(S) \coloneq \bE_\xi[f_q(S,\xi) | S]\quad \text{and}\quad\bE[\varepsilon_q(S,\xi) | S] = 0,\label{eq:observed_readouts}
\end{equation}
where $\varepsilon_q(S,\xi)$ is the noise arising from the training randomness $\xi$, e.g., data ordering or random initialization. Note, that a single training run is shared by all queries, so the noises $\varepsilon_q$ are correlated across $q$. It is common practice to use $\bar{f}_{R,q}(S) = F_q(S) + \tfrac1R\sum_{r=1}^R \varepsilon_q(S, \xi_r)$ as a proxy for $F_q(S)$, with, for example, $R=5$ \citep[Appendix~C.4]{Park2023TRAK} or higher \citep[Appendix~B.2]{Bae2024SOURCE}. 

\paragraph{Attribution Method Noise.} The attribution predictions are also stochastic. The computed scores $g_q(S,\zeta)$ depend on the TDA method's own randomness $\zeta$ (e.g.\ randomness of the trained model, random projections in TRAK \citep{Park2023TRAK}, MC sampling in EK-FAC \citep{Grosse2023EKFAC}). Namely we have
\begin{equation}
    g_q(S, \zeta) = G_q(S) + \gamma_q(S, \zeta),\qquad \text{with}\quad \bE_\zeta[g_q(S,\zeta) | S] = G_q(S)\quad \text{and}\quad \bE[\gamma_q(S,\zeta) | S] = 0,\label{eq:observed_attributions}
\end{equation}
where $\zeta$ is the randomness of the attribution method resulting in the TDA noise $\gamma_q(S,\zeta)$. As a result, it is a common practice to compute $E$ independently random instances of the same TDA method and average their attribution scores $\bar{g}_{E,q}(S) = G_q(S) + \frac{1}{E}\sum_{e=1}^E\gamma_q(S,\zeta_{e})$ to reduce the effect of the noise. 

\paragraph{Finite-Subset Error.} In practice, using a finite number $M$ of subsets introduces an error between $\empsp_M$ and $\spearman$. To control the error, naive LDS estimation requires a large $M$. When combining all finite-sample effects, we end up with the following estimate
\[
\empsp_{R,E,M}(q) = \Corr_{S \sim \hatpiS}(\rank_M(\bar{f}_{R,q}(S)),\rank_M(\bar{g}_{E,q}(S))).
\]

\section{Improved LDS Estimation}\label{sec:corrected}
We first estimate LDS against expected retrained outputs while holding a finite attribution ensemble fixed. We then consider two related targets: the LDS of a conventional finite-seed evaluation and the training-noise-free LDS of a larger attribution ensemble. Finally, we address the bias from estimating Spearman correlation using finitely many subsets.

\subsection{Correcting for the Variance of the Training and the Attribution Noise}

We show how estimating the variances of training and attribution noise can reduce the cost of LDS evaluation.

In the following we assume the randomness of the training and the TDA method are independent.
\begin{assumption}[Independent Training Seeds]\label{asm:noise}
For each subset $S$, the $R$ retraining seeds $\xi_1,\dots,\xi_R$ are i.i.d.\ and are further independent of the $E$ i.i.d.\ ensemble seeds $\zeta_1,\dots,\zeta_E$.
\end{assumption}
In practice, it is always possible to design such an experiment by using different random seeds for the training and the TDA method and thus this assumption is not restrictive. 

\paragraph{Correcting training noise.}
For a single query, we omit the index $q$ to ease the notation and consider Pearson LDS computed from $M$ subsets, $R$ training seeds per subset, and a TDA ensemble with $E$ members:
\begin{equation}\label{eq:pearson_lds_empirical}
    \emppe_{R,E,M} = \frac{\Cov_{S\sim\hatpiS}(\bar{f}_{R}(S),\bar{g}_{E}(S))}{\sqrt{\Var_{S \sim \hatpiS}(\bar{f}_{R}(S))\Var_{S \sim \hatpiS}(\bar{g}_E(S))}}.
\end{equation}

To isolate the effect of training noise, fix the attribution
ensemble $\bar g_E$, and consider the \emph{population correlation} over the training subsets $S\sim\piS$ and retraining randomness. Under \Cref{asm:noise}, the zero-mean training noise leaves the covariance unchanged, while averaging independent retraining seeds reduces its variance:
\[
    \Cov_{S,\xi_{1:R}}(\bar g_E,\bar f_R)=\Cov_S(\bar g_E,F), \qquad \Var_{S,\xi_{1:R}}(\bar f_R)=\sigma_F^2+\frac{\sigma_\varepsilon^2}{R},
\]
where $\sigma_F^2=\Var_S(F)$ and $\sigma_\varepsilon^2=\Var_{S,\xi}(\varepsilon)$. Writing $\lambda=\sigma_\varepsilon^2/\sigma_F^2$ for the training-noise to signal ratio, we obtain
\begin{equation}\label{eq:training-attenuation}
    \rho^{\mathrm p}_{R,E}
    \coloneqq
    \Corr_{S,\xi_{1:R}}(\bar g_E(S),\bar f_R(S))
    =
    \frac{\rho^{\mathrm p}_E}
         {\sqrt{1+\lambda/R}},
\end{equation}
where $\rho^{\mathrm p}_{E}=\Corr_S(\bar g_E(S),F(S))$ is the population LDS without retraining noise, for the same attribution ensemble. Equation~\eqref{eq:training-attenuation} describes classical \emph{noise attenuation} due to measurement error in one variable \citep[Part~II, \S2, pp.~89--90]{Spearman1904}. Since standard LDS evaluations use a fixed finite number of retrainings per subset \citep{Park2023TRAK,Georgiev2023Journey,Bae2024SOURCE, Deng2024Dattri,Deng2024Ensembles}, their Pearson counterpart estimates the attenuated correlation $\rho^{\mathrm p}_{R,E}$ rather than $\rho^{\mathrm p}_E$, even as $M\to\infty$. Under the joint Gaussian assumptions used in \eqref{eq:pearson-spearman}, the standard Spearman LDS is likewise attenuated towards zero. 

Independent retrainings allow us to estimate the noise variance and subtract its contribution to the observed readout variance. \Cref{alg:noise-corrected-lds} computes the sample variance across retraining seeds for each subset, then averages these variances to obtain $\widehat{\sigma}_\varepsilon^2$. The corrected estimated signal variance is therefore $\widehat{\sigma}_F^2 =\widehat{\Var}_M(\bar f_R)-\widehat{\sigma}_\varepsilon^2/R$. This applies the classical correction for attenuation proposed by \citet{Spearman1904} to the retraining outputs, while holding the attribution predictor fixed.

At a fixed budget of the number of retrains $N=MR$, naive averaging with more seeds, i.e., larger $R$, reduces the training noise bias, but leaves fewer subsets for estimating the correlation, which the following proposition makes precise.
\begin{wrapfigure}{R}{0.43\textwidth}
  \vspace{-1\intextsep}
  \begin{minipage}{\linewidth}
  \begin{algorithm}[H]
  \small
  \caption{$R$-seed noise-corrected LDS}
  \label{alg:noise-corrected-lds}
  \begin{algorithmic}
  \Require $\bar g_E$; subsets $S_1,\dots,S_M$; $M,R\geq2$
  \For{$m = 1$ \textbf{to} $M$}
  \For{$r = 1$ \textbf{to} $R$}
  \State $f_{m,r}\gets f(S_m, \xi_{m,r})$
  \EndFor
  \State $\bar{f}_m\gets \tfrac1R\sum_{r} f_{m,r}$
  \State $s^2_m\gets \tfrac{1}{R-1}\sum_{r} (f_{m,r}-\bar{f}_m)^2$
  \EndFor
  \State $\widehat{\sigma}_\varepsilon^2\gets \tfrac1M\sum_{m} s^2_m$
  \State $\widehat{\sigma}_f^2\gets \widehat{\Var}_M(\bar{f}_m)$
  \State $\widehat{\sigma}_F^2\gets \widehat{\sigma}_f^2-\widehat{\sigma}_\varepsilon^2/R$
  \State $\widehat{\sigma}_g^2\gets \widehat{\Var}_M(\bar g_E(S_m))$
  \State $\widehat{C}\gets \widehat{\Cov}_M(\bar{f}_m,\,\bar g_E(S_m))$
  \If{$\widehat{\sigma}_F^2\leq 0$ or $\widehat{\sigma}_g^2\leq 0$}
  \State \Return $0$
  \EndIf
  \State \Return $\operatorname{clip}_{[-1,1]}\!\left(\widehat{C}/(\widehat{\sigma}_g\,\widehat{\sigma}_F)\right)$
  \end{algorithmic}
  \end{algorithm}
  \end{minipage}
  \vspace{-3\intextsep}
\end{wrapfigure}

\begin{proposition}[RMSE of naive seed averaging]\label{prop:naive}
Fix the attribution ensemble $\bar g_E$. Suppose \cref{asm:noise} holds, the observations are i.i.d.\ across subsets, $\bar g_E,F,\varepsilon$ have finite fourth moments, and $\Var_S(\bar g_E),\Var_S(F)>0$.\footnote{See \cref{app:complexity} for the full assumptions, estimator definitions, and finite-sample bounds.} Then the sample Pearson correlation $\emppe_{R,E,M}$ of $(\bar g_E(S_m),\bar f_R(S_m))_{m=1}^M$ satisfies
\begin{equation*}
    \bE\bigl[(\emppe_{R,E,M}-\rho^{\mathrm p}_E)^2\bigr]^{1/2} =\bigO\!\left(\sqrt{\frac{R}{N}}+\frac{\lambda}{R}\right),
\end{equation*}
with a constant that is independent of $M\geq2$ and $R\geq1$. For fixed $\lambda>0$, allocating $R\asymp\lambda^{2/3}N^{1/3}$ gives RMSE $\bigO(\lambda^{1/3}N^{-1/3})$ as $N\to\infty$.
\end{proposition}
Balancing the sampling and the noise attenuation terms in Proposition~\ref{prop:naive} gives the $\bigO(N^{-1/3})$ bound for naive averaging; the proof is given in \cref{subsec:proof_prop_naive}. However, the variance correction in \Cref{alg:noise-corrected-lds} achieves the following $\bigO(N^{-1/2})$ bound with a fixed number of seeds $R\geq 2$.

\begin{proposition}[RMSE of training-noise correction]\label{prop:corrected}
Let $\unbpe_{R,E,M}$ denote the estimate returned by \cref{alg:noise-corrected-lds} with the fixed attribution predictor $\bar g_E$. Under the assumptions of Proposition~\ref{prop:naive}, for any fixed number of seeds $R\geq2$,
\begin{equation}\label{eq:corrected-rmse}
\bE\bigl[(\unbpe_{R,E,M}-\rho^{\mathrm p}_E)^2\bigr]^{1/2}
=\bigO(N^{-1/2})
\end{equation}
as $N=MR\to\infty$ and the constant can depend on $R$.
\end{proposition}
See the proof in \cref{subsec:proof_prop_corrected}. 

\paragraph{Reconstructing finite-seed LDS from fewer seeds.}
To compare with existing evaluations, we may wish to reproduce the LDS targeted by a naive $R$-seed averaging protocol, which includes the attenuation from the training noise. By estimating the noise-free Pearson LDS and the noise-to-signal ratio $\lambda$ from as few as $k\geq2$ seeds per subset, we can then reconstruct the finite-seed target $\rho^{\mathrm p}_{R,E}$ as

\begin{equation}\label{eq:finite-seed-reconstruction}
    \widehat{\rho}^{\,\mathrm{p,rec}}_{R,E,M}
    =
    \frac{\unbpe_{k,E,M}}
         {\sqrt{1+\widehat{\lambda}/R}}.
\end{equation}

For $k<R$, this uses fewer retrainings than directly averaging $R$ seeds per subset. Estimator details are given in \cref{app:finite-seed-reconstruction}.

\begin{figure}[t]
    \centering
    \includegraphics[width=1\linewidth]{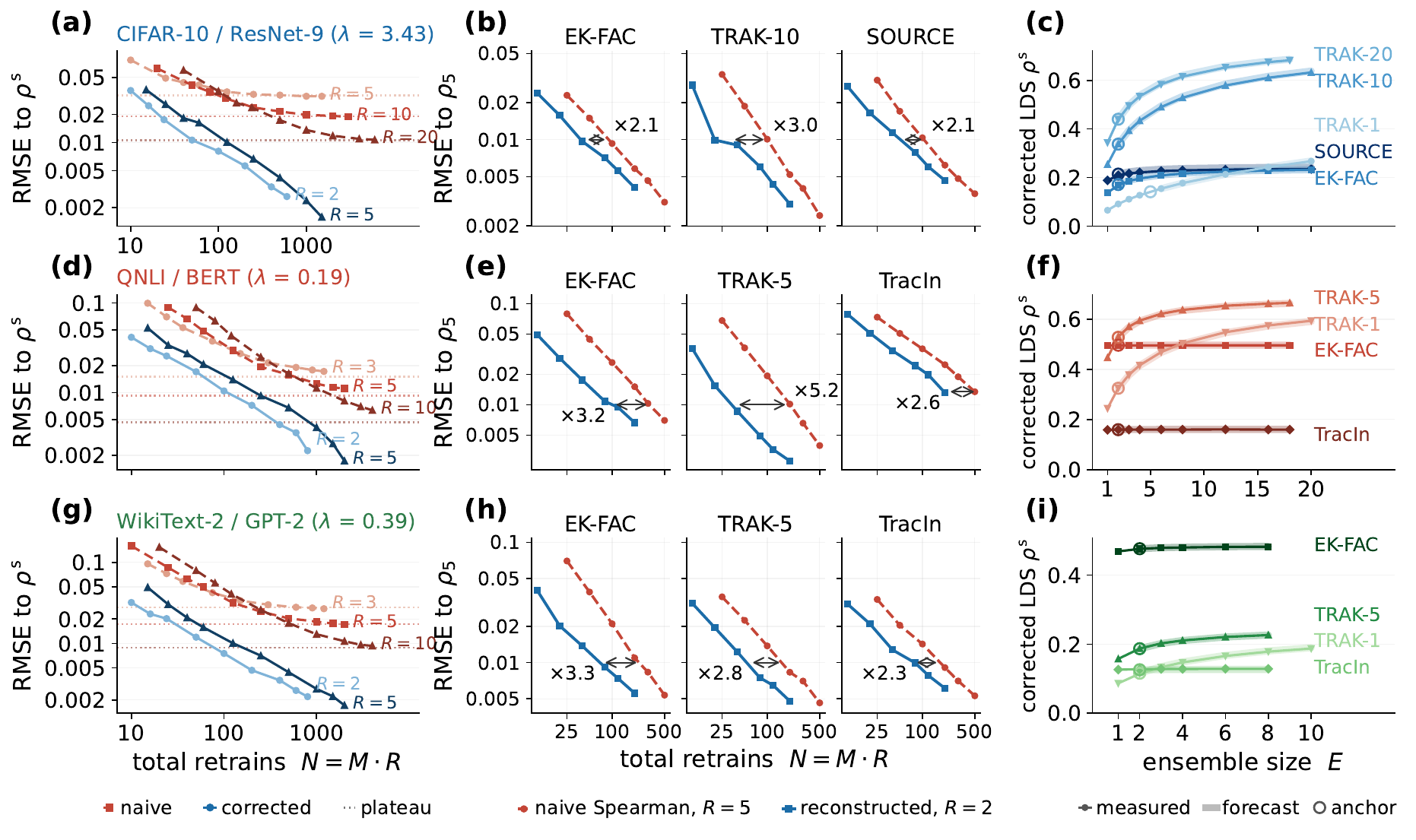}
    \caption{\textbf{Noise correction reduces the cost of LDS evaluation.} Rows: CIFAR-10/ResNet-9, QNLI/BERT, WikiText-2/GPT-2. \textbf{(a, d, g)} RMSE to the noise-free Spearman LDS versus total retrainings, for naive $R$-seed averaging and the noise correction (EK-FAC); dotted: predicted naive plateaus. \textbf{(b, e, h)} Reconstructing the five-seed Spearman target $\rho_5$ from two seeds matches naive averaging with $2.1$--$5.2\times$ fewer retrainings. \textbf{(c, f, i)} Measured noise-corrected Spearman LDS and predictions from the circled small ensembles. The first two columns use $E=1$ and disjoint retraining seeds for estimation and reference.}
    \label{fig:correction}
\end{figure}

\paragraph{Predicting LDS of TDA ensembles.}
Attribution ensembling has the same $1/E$ noise-variance scaling. This motivates a forecast analogous to \eqref{eq:training-attenuation}, with $(R,\lambda)$ replaced
by $(E,c)$, where $c=\bE_\zeta[\Var_S(\gamma(S,\zeta))]/\Var_S(G)$ is the attribution noise-to-signal ratio\footnote{This is the analogue to the previously defined $\lambda$, but here on the attribution arm.}. Using $R$ retraining seeds on $M$ subsets, we can forecast the training-noise-free LDS of an $E$-member ensemble from $K \leq E$ observed attribution members as
\begin{equation}\label{eq:finite-ensemble-prediction}
    \widehat{\rho}^{\,\mathrm{p,pred}}_{R,E,M}
    \coloneqq
    \unbpe_{R,K,M}
    \sqrt{\frac{1+\widehat c/K}{1+\widehat c/E}},
    \qquad E \geq K\geq2.
\end{equation}
Here $\unbpe_{R,K,M}$ is the training-noise-corrected Pearson LDS of the observed $K$-member ensemble. We estimate $c$ from the variance across subsets of differences between independent attribution members; see \cref{app:ensemble-correction}. Thus, a small attribution ensemble can forecast the LDS of larger ensembles, as tested in \cref{fig:fig1b,fig:correction}.

\begin{wrapfigure}{r}{0.4\textwidth}
    \vspace{-\intextsep}
    \centering
    \captionsetup{font=footnotesize,justification=raggedright,singlelinecheck=false}
    \includegraphics[width=\linewidth]{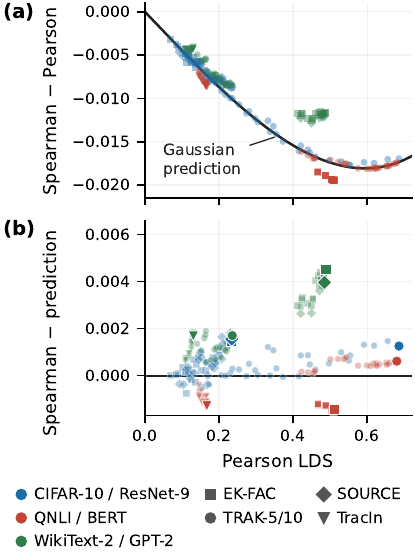}
    \caption{\textbf{Spearman versus Pearson LDS.} \textbf{(a)}~Spearman minus Pearson LDS; curve: Gaussian prediction. \textbf{(b)}~Residual from the finite-$M$ Gaussian expectation. Large markers: full ensemble, all seeds.}
    \label{fig:arcsin-bridge}
    \vspace{-1\intextsep}
\end{wrapfigure}

\paragraph{Converting Pearson LDS to Spearman LDS.}
To express the corrections above in terms of Spearman LDS, we use a Gaussian assumption.
\begin{assumption}[Gaussian signal pair]\label{asm:gauss}
The pair $\left(G(S), F(S)\right)$ is jointly Gaussian under $S\sim\piS$.
\end{assumption}
Under \cref{asm:gauss}, the population Spearman and Pearson correlations satisfy
\begin{equation}
    \spearman(q) = \frac{6}{\pi}\arcsin(\pearson(q) / 2).\label{eq:pearson-spearman}
\end{equation}
For finite-$R$ or finite-$E$ targets, the assumption requires joint Gaussianity of the corresponding averaged prediction-readout pair. We show in \cref{fig:arcsin-bridge} that average discrapancies from the Gaussian prediction remain below $0.006$ across $220$ of the tested settings. In \cref{fig:qq-normality} in the appendices we see seed averaging with larger $R$ improves agreement with the Gaussian reference on CIFAR-10 while we find departures from Gaussianity persist on QNLI and WikiText-2.

\Needspace{24\baselineskip}

\paragraph{Numerical tests.} \Cref{fig:correction} tests the corrections on ResNet-9/CIFAR-10, BERT/QNLI and GPT-2/WikiText-2, using TRAK, EK-FAC \citep{Grosse2023EKFAC}, SOURCE \citep{Bae2024SOURCE} and TracIn \citep{Pruthi2020TracIn}. Taking the median over queries on the full banks gives estimated retraining
noise-to-signal ratios of $3.43$, $0.19$ and $0.39$ for CIFAR-10, QNLI and WikiText-2, respectively. We specify the experimental setup in \cref{app:experimental-details}.

The first column of \cref{fig:correction}(a,~d,~g) plots the RMSE of naive and corrected Spearman LDS against total retrainings, relative to noise-corrected reference scores computed from held-out retraining seeds. The naive estimates obtained by averaging $R$ seeds stall at the plateaus predicted by the attenuation \eqref{eq:training-attenuation}, while the corrected estimates keep improving with the budget. On CIFAR-10, which has the largest noise-to-signal ratio, the two-seed correction substantially reduces the bias that remains even with naive $R=20$ averaging. \Cref{fig:fig6-rmse,fig:fig6-single-draw} further illustrate the effect of correcting training noise when targeting the noise-free LDS, showing RMSE across repeated subsampling draws and estimates from a single draw, respectively.

The second column of \cref{fig:correction}(b,~e,~h) plots RMSE against total retrainings, using a reference for the five-seed Spearman target $\rho_5$ computed from held-out retraining seeds. Reconstructing this target from $k=2$ seeds per subset, using the Pearson--Spearman conversion, matches naive five-seed averaging with $2$--$5\times$ fewer retrainings ($2.3$--$3.3\times$ on WikiText-2). Thus, even the biased finite-seed LDS can be estimated more efficiently via \eqref{eq:finite-seed-reconstruction}.

The third column of \cref{fig:correction}(c,~f,~i) shows that small ensembles can accurately predict the LDS of larger ensembles using the prediction formula in \eqref{eq:finite-ensemble-prediction}.\footnote{For TRAK-$k$, each member uses $k$ models for featurization \citep{Park2023TRAK}, so $E$ members use $kE$ models.} Forecasts from just $K=2$ members ($K=5$ for TRAK-1 on CIFAR-10) are within $0.008$ of measured noise-corrected Spearman LDS on CIFAR-10 and QNLI, and within $0.005$ on WikiText-2 over the plotted ensemble sizes. TRAK-1 attribution scores have higher variance and therefore require more members to estimate the noise-to-signal ratio $c$ reliably, which is why we base their CIFAR-10 forecast at $K=5$ rather than $K=2$ (\cref{app:experimental-details}).

\Cref{fig:arcsin-bridge} compares Pearson and Spearman LDS across 220 settings, varying the dataset, attribution method, $E$ and $R$. Their difference reaches about $0.02$ in panel (a), while residuals from the exact finite-$M$ Gaussian expectation in panel (b) remain below $0.006$, with the largest departures on WikiText-2. This supports the conversion as an approximation in the tested settings and we give tests of Gaussianity in \cref{fig:qq-normality} in the appendix.

\par\ifnum\value{WF@wrappedlines}>1
    \vspace{\dimexpr\value{WF@wrappedlines}\baselineskip-\baselineskip\relax}
\fi
\WFclear
\Needspace{0.45\textheight}
\subsection{Finite-Subset Bias}\label{sec:finite-subset}
\begin{wrapfigure}{r}{0.4\textwidth}
    \centering
    \vspace{-1\intextsep}
    \includegraphics[width=\linewidth]{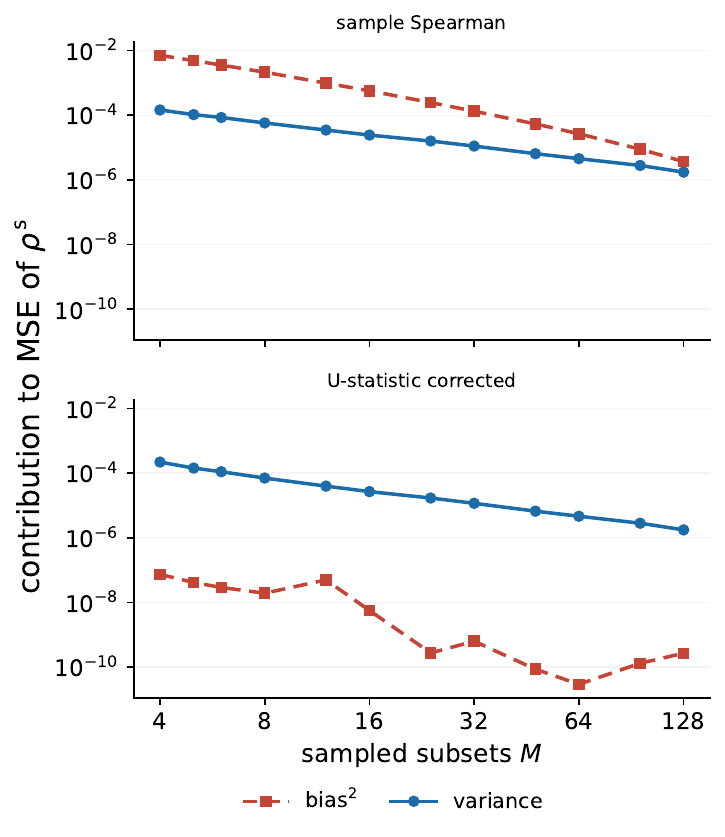}
    \caption{Bias (squared) and variance of estimating TRAK LDS on CIFAR-10 using sample Spearman $\empsp$ and the unbiased estimator $\unbsp$. Using the unbiased estimator removes the large bias in estimating LDS.}
    \label{fig:bias_variance}
    \vspace{-1\intextsep}
\end{wrapfigure}
We now address the bias from estimating Spearman correlation using finitely many subsets. This bias is particularly relevant when LDS is averaged over many queries, because, while the effect of averaging can reduce sampling variance, the systematic bias persists. In our experiments, bias dominates the estimation error (\Cref{fig:bias_variance,fig:bias_variance_queries}). This motivates using an unbiased $U$-statistic estimator \citep{hoeffding1948class}, despite its potential variance cost.

Let $\hat\tau_M$ denote Kendall's sample tau-a, averaged over the same queries as $\empsp_M$. For $M\geq3$ i.i.d.\ observations with continuous marginals (and hence no ties),
\begin{equation}\label{eq:sp_unb}
        \unbsp_M \coloneqq \frac{(M+1)\empsp_M - 3\hat{\tau}_M}{M - 2}
\end{equation}
is an unbiased estimator of $\spearman$, i.e.\ $\E{\unbsp_M} = \spearman$~\cite[\S 9]{hoeffding1948class}. The definition of $\hat\tau_M$ and the proof are given in \cref{app:various_correlations}. 

In \cref{fig:fig1c}, subsets are sampled without replacement from a precomputed retraining bank containing model readouts for $300$ training subsets, with $40$ retraining seeds per subset. We use the finite-population adjustment in \eqref{eq:finite-bank-spearman} to target the Spearman LDS computed on all $300$ subsets. \Cref{fig:bias_variance} shows how bias is the dominant term in estimating the Spearman correlation for the CIFAR-10 dataset, and how using \eqref{eq:sp_unb} removes this bias. We test this debiasing for three different TDA methods, namely TRAK, EK-FAC, and SOURCE, on CIFAR-10 and MNIST datasets, where for all instances the model is trained from scratch. The result in \Cref{fig:spearman_grid} shows similar gains in reducing the estimation error. Averaging over queries reduces variance, making finite-subset bias the dominant source of error in these experiments. This is demonstrated in \Cref{fig:bias_variance_queries} in the appendix.

\section{Discussion}\label{sec:discussion}
Our results show that LDS evaluation can be made substantially cheaper by correcting for training noise and finite-subset bias, and by predicting ensemble performance from a few attribution models. The gains depend on the noise level and on whether the target is a biased finite LDS or the score without training noise.
\section*{Acknowledgments}
This work was conducted as part of the ML Alignment \& Theory Scholars (MATS) Program, with support from MATS. We thank Ryan Carey for feedback and helpful discussions.

\bibliographystyle{plainnat}
\bibliography{references}

\newpage
\appendix
\raggedbottom
\section{Proofs of the estimation error bounds}\label{app:complexity}

\paragraph{Sampling and estimator conventions.}
We analyse one query, conditional on the attribution seeds, which are sampled independently of the evaluation subsets and retraining seeds. This fixes the attribution predictor $\bar g_E$. Throughout these two proofs, we write $G=\bar g_E$ and $\pearson=\rho_E^{\mathrm p}=\Corr_{S\sim\piS}(G(S),F(S))$, suppressing the ensemble index $E$. All expectations below are conditional on the attribution seeds and average over evaluation subsets and retraining randomness.

The observations for each subset $(S_m,f_{m,1},\dots,f_{m,R})$, $m=1,\dots,M$, are i.i.d., with $S_m\sim\piS$. Conditional on each subset, the $R$ readouts are i.i.d.\ with mean $F(S)$ and residual $\varepsilon$ as in \eqref{eq:observed_readouts}. Write
$\sigma_G^2=\Var_S(G)>0$, $\sigma_F^2=\Var_S(F)>0$,
$\sigma_\varepsilon^2=\bE[\varepsilon^2]$,
$\lambda=\sigma_\varepsilon^2/\sigma_F^2$, and $N=MR$.
Assume finite fourth moments, and define
\[
\kappa \coloneqq \max\left\{
 \frac{\bE[(G-\bE G)^4]}{\sigma_G^4},\,
 \frac{\bE[(F-\bE F)^4]}{\sigma_F^4},\,
 \frac{\bE[\varepsilon^4]}{\sigma_\varepsilon^4}
\right\}<\infty,
\]
where we define the last ratio as zero when $\sigma_\varepsilon^2=0$. 

Across-subset sample variances and covariances use divisor $M-1$; sample variances across the $R$ retrainings of each subset use divisor $R-1$. We set the sample correlation to zero whenever either sample variance vanishes.

\subsection{Proof of Proposition~\ref{prop:naive}}\label{subsec:proof_prop_naive}
\begin{proof}
Let $\bar f_R(S)=R^{-1}\sum_{r=1}^R f(S,\xi_r)$ be the model readout averaged over $R$ retrainings of subset $S$. Define the standardised attribution prediction $X$ and mean model response $Z$, together with the scaled retraining noise and averaged readout:
\[
X=\frac{G-\bE G}{\sigma_G},\qquad
Z=\frac{F-\bE F}{\sigma_F},\qquad
\eta_r=\frac{\varepsilon(S,\xi_r)}{\sigma_F},\qquad
Y_R=Z+\bar\eta_R,\quad
\bar\eta_R=\frac1R\sum_{r=1}^R\eta_r.
\]
Here $Y_R=(\bar f_R-\bE F)/\sigma_F$ is the centred averaged readout scaled by the signal standard deviation. Then $\Var(X)=\Var(Z)=1$, $\bE[\eta_r^2]=\lambda$, and
$\Cov(X,Y_R)=\pearson$. Conditional independence and zero conditional means give
\begin{align*}
\bE[\bar\eta_R^4]
&=\frac{\bE[\eta_1^4]}{R^3}
 +\frac{3(R-1)}{R^3}
   \bE\!\left[\bE[\eta_1^2\mid S]^2\right]\leq \frac{3\kappa\lambda^2}{R^2}.
\end{align*}
Writing $t=\lambda/R$, it follows that
$\Var(Y_R)=1+t$ and
\[
\bE[Y_R^4]\leq 8\bE[Z^4]+8\bE[\bar\eta_R^4]
\leq24\kappa(1+t)^2.
\]
Now standardise the averaged readout by its own standard deviation:
\[
V_R=\frac{Y_R}{\sqrt{1+t}}
=\frac{\bar f_R-\bE F}{\sqrt{\sigma_F^2+\sigma_\varepsilon^2/R}}.
\]
Thus the standardised attribution prediction $X$ and standardised averaged readout $V_R$ both have unit variance and fourth moments at most $24\kappa$, uniformly in $R$. Their sample correlation equals $\emppe_{R,M}$, since centring and positive rescaling do not change Pearson correlation.

For $M$ i.i.d.\ copies of centred variables $U,V$ with covariance $c$, the sample covariance $\widehat c$ satisfies
\begin{equation}\label{eq:sample-moment-bound}
\bE[(\widehat c-c)^2]
=\frac{\bE[U^2V^2]-c^2}{M}
 +\frac{\Var(U)\Var(V)+c^2}{M(M-1)}
\leq\frac{3\sqrt{\bE[U^4]\bE[V^4]}}{M}.
\end{equation}
Setting $U=V$ gives $\bE[(\widehat{\Var}_M(U)-\Var(U))^2]\leq 3\bE[U^4]/M$ for the mean squared error of the sample variance. Applying these bounds to the $M$ observed pairs $(X_m,V_{R,m})$, one per evaluation subset, shows that the sum of the mean squared errors of the two across-subset sample variances and their sample covariance is at most $C_0\kappa/M$, for a universal constant $C_0$.

On the event that the sample variances of $X$ and $V_R$ are both at least $1/2$, the correlation error is bounded by a universal constant times the sum of these three moment errors. To see this, use the Lipschitz continuity of $(a,b)\mapsto (ab)^{-1/2}$ on $a,b\geq1/2$ and split off the covariance error. By Chebyshev's inequality, the complementary event has probability at most $C_1\kappa/M$; there the error is at most $2$, including the specified zero convention. Hence
\[
\bE[(\emppe_{R,M}-\rho_R^{\mathrm p})^2]^{1/2}
\leq C\sqrt{\kappa/M}.
\]
The triangle inequality in $L^2$, \eqref{eq:training-attenuation}, and
$1-(1+x)^{-1/2}\leq x/2$ for $x\geq0$ therefore give
\begin{equation}\label{eq:naive-rmse-explicit}
\bE\bigl[(\emppe_{R,M}-\pearson)^2\bigr]^{1/2}
\leq C\sqrt{\kappa}\sqrt{\frac{R}{N}}
 +\frac{|\pearson|\lambda}{2R},
\end{equation}
where $C$ is universal and $N=MR$. For fixed $\kappa$ and $\lambda>0$, taking $R\asymp\lambda^{2/3}N^{1/3}$ gives RMSE $\bigO(\lambda^{1/3}N^{-1/3})$, as stated in the proposition.
\end{proof}

\subsection{Proof of Proposition~\ref{prop:corrected}}\label{subsec:proof_prop_corrected}
\begin{proof}
Use the normalised variables $X,Y_R,\eta_r$ and $t=\lambda/R$ from the preceding proof. Let
\[
\widehat a=\widehat{\Var}_M(X),\qquad
\widehat c=\widehat{\Cov}_M(X,Y_R),\qquad
\widehat w=\frac1M\sum_{m=1}^M s_{\eta,m}^2,\qquad
\widehat d=\widehat{\Var}_M(Y_R)-\widehat w/R,
\]
where $s_{\eta,m}^2$ is the within subset sample variance of the $R$ noise replicates,
\[
s_{\eta,m}^2\coloneqq\frac{1}{R-1}\sum_{r=1}^R(\eta_{m,r}-\bar\eta_m)^2,
\qquad
\bar\eta_m\coloneqq\frac1R\sum_{r=1}^R\eta_{m,r},
\]
with $\eta_{m,r}=\varepsilon(S_m,\xi_{m,r})/\sigma_F$. The normalised readouts $f_{m,r}/\sigma_F$ have the same within subset sample variance, since their shared signal cancels. Thus
$\bE[\widehat a]=\bE[\widehat d]=1$,
$\bE[\widehat c]=\pearson$, and $\bE[\widehat w]=\lambda$.

Expanding the centred sum of squares gives
$s_{\eta,m}^2=(R-1)^{-1}(\sum_r\eta_{m,r}^2-R\bar\eta_m^2)\leq(R-1)^{-1}\sum_r\eta_{m,r}^2$.
By Cauchy--Schwarz, $(\sum_r\eta_{m,r}^2)^2\leq R\sum_r\eta_{m,r}^4$.
Thus, for $R\geq2$, the second moment of the sample variance satisfies
$\bE[(s_{\eta,m}^2)^2]\leq (R/(R-1))^2\bE[\eta_1^4]\leq4\kappa\lambda^2$.
Independence across subsets therefore implies
\[
\bE[(\widehat w/R-t)^2]\leq\frac{4\kappa t^2}{M}.
\]
Combining this with the sample-moment bound in \eqref{eq:sample-moment-bound}, including its $U=V$ case, yields
\[
\bE[(\widehat a-1)^2]
+\bE[(\widehat c-\pearson)^2]
+\bE[(\widehat d-1)^2]
\leq \frac{C_2\kappa(1+t)^2}{M}.
\]
This step uses the inequality $(u-v)^2\leq2u^2+2v^2$ and does not require independence between $\widehat w$ and the across-subset sample variance.

On $\{\widehat a\geq1/2,\widehat d\geq1/2\}$, the same ratio bound as above controls the error of $\widehat c/\sqrt{\widehat a\widehat d}$. Projection onto $[-1,1]$ cannot increase its distance to $\pearson$. On the complementary event, the returned estimate is still in $[-1,1]$, so its squared error is at most $4$. Chebyshev's inequality consequently gives
\begin{equation}\label{eq:corrected-rmse-explicit}
\bE\bigl[(\unbpe_{R,M}-\pearson)^2\bigr]^{1/2}
\leq C\sqrt{\kappa}\left(1+\frac{\lambda}{R}\right)\sqrt{\frac{R}{N}},
\end{equation}
where we substituted $t=\lambda/R$ and $M=N/R$. For fixed $R\geq2$ and a fixed data-generating distribution, this is $\bigO(N^{-1/2})$, proving \eqref{eq:corrected-rmse}.
\end{proof}

\section{Derivation of the finite-subset Spearman correction}\label{app:various_correlations}
In this section we provide a derivation of the classical unbiased Spearman estimator \citep{hoeffding1948class} used in \cref{sec:finite-subset}. We then adapt it to the finite retraining-bank reference used in \cref{fig:fig1c}.

Fix a query, and let $(X,Y)$ denote the attribution prediction and model readout for a randomly drawn training subset. Consider $m \geq 3$ independent observations from this joint distribution. We assume $X,Y$ have continuous marginals, so that ties occur with probability zero. Writing $F_X$ and $F_Y$ for the marginal distribution functions, the population Spearman and Kendall correlations are
\begin{equation*}
  \spearman = 12 \, \E{F_X(X) F_Y(Y)}-3 \qquad\text{and}\qquad \tau = \E{\sign(X_1 - X_2)\sign(Y_1 - Y_2)},
\end{equation*}
where the two pairs are independent copies of $(X,Y)$.

When given $m$ i.i.d.\ observations $\{X_i,Y_i\}_{i=1}^m$, let $R_i \coloneqq 1 + \sum_{j \neq i} 1(X_j < X_i)$ denote the rank of $X_i$ (smallest first), and similarly $S_i \coloneqq 1 + \sum_{j \neq i} 1(Y_j < Y_i)$. 

\paragraph{Finite-sample bias.}
The finite-sample Spearman and Kendall coefficients are given by
\[
\empsp = \frac{12}{m(m^2-1)}\sum_{i=1}^m R_iS_i - \frac{3(m+1)}{m-1}
\]
and
\[
\hat{\tau} = \frac{1}{{m \choose 2}}\sum_{j < i}\sign(X_i - X_j)\sign(Y_i - Y_j)
\]
respectively. To compute the expectation of sample Spearman, we expand $R_i S_i$. The constant and single-indicator terms contribute $m$ in expectation. Among the remaining terms, $m-1$ compare both coordinates against the same observation, while $(m-1)(m-2)$ use two distinct comparison observations:
\begin{align*}
    \E{R_iS_i} &= m + (m-1)\Parg{X_1 > X_2, Y_1 > Y_2} + (m-1)(m-2)\E{F_X(X_1)F_Y(Y_1)}\\
    &= m + (m-1)\cdot \frac{\tau + 1}{4} + (m-1)(m-2)\cdot \frac{\spearman + 3}{12},
\end{align*}
where the same observation terms have probability \((1+\tau)/4\). For distinct comparison observations, conditioning on \((X_i,Y_i)\) gives the product \(F_X(X_i)F_Y(Y_i)\), whose expectation is \((\spearman+3)/12\).

By plugging this back into the equation for $\empsp$, we obtain
\begin{equation}
    \E{\empsp} = \frac{(m-2)\spearman + 3\tau}{m+1}.
\end{equation}
This implies that the finite-sample $\empsp$ has a bias of $3(\tau - \spearman) / (m+1)$. 

\paragraph{An unbiased estimator.}
Each pairwise term in sample Kendall's tau has expectation $\tau$, so $\E{\hat{\tau}}=\tau$. Consequently, the estimator
\begin{equation}
    \unbsp \coloneqq \frac{(m+1)\empsp - 3\hat{\tau}}{m - 2}
\end{equation}
is unbiased for $m \geq 3$, i.e.\ $\E{\unbsp} = \spearman$. This is the estimator in \eqref{eq:sp_unb}. By linearity of expectation, averaging the corrected estimates over queries gives an unbiased estimate of the query-averaged population Spearman correlation.

\paragraph{Adjustment for a finite retraining bank.}
For \cref{fig:fig1c}, we sample $m$ prediction/readout pairs uniformly without replacement from a fixed collection of $P=300$ pairs, each corresponding to a training subset. There are no ties in either marginal for any query. The target is the ordinary Spearman correlation $\empsp_P$ computed on all $P$ pairs. The corresponding correction is
\begin{equation}\label{eq:finite-bank-spearman}
\widehat\rho^{\,\mathrm{s,bank}}_{m,P}
\coloneqq \frac{(P-2)\unbsp_m+3\hat\tau_m}{P+1},
\qquad 3\leq m\leq P.
\end{equation}
Here, expectations are taken over the subsample draw, conditional on the fixed bank. Since $\unbsp_m$ and $\hat\tau_m$ are U-statistics of orders three and two, uniform sampling without replacement gives $\E{\unbsp_m}=\unbsp_P$ and $\E{\hat\tau_m}=\hat\tau_P$. Substituting the statistics computed on all $P$ pairs into \eqref{eq:finite-bank-spearman} gives $\E{\widehat\rho^{\,\mathrm{s,bank}}_{m,P}}=\empsp_P$. The same identities hold after averaging over queries.

\section{Estimator details}
\subsection{Reconstructing a finite-seed LDS}\label{app:finite-seed-reconstruction}
Let $\bar f_k$ average $k\geq2$ retrainings per subset, and estimate $\widehat{\sigma}_\varepsilon^2$ as in \cref{alg:noise-corrected-lds}. The observed readout variance contains $\sigma_\varepsilon^2/k$ of training noise, whereas the target $R$-seed protocol contains $\sigma_\varepsilon^2/R$. Subtracting their estimated difference gives
\[
\widehat{\rho}^{\,\mathrm{p,rec}}_R
=\frac{\widehat{\Cov}_M(G,\bar f_k)}
{\sqrt{\widehat{\Var}_M(G)
\left(\widehat{\Var}_M(\bar f_k)
-(1/k-1/R)\widehat{\sigma}_\varepsilon^2\right)}}.
\]
When the estimated signal and predictor variances are positive, this is \eqref{eq:finite-seed-reconstruction} with $\widehat\lambda=\widehat{\sigma}_\varepsilon^2/\widehat{\sigma}_F^2$, before variance safeguards and final clipping. At $R=k$ it reduces to the ordinary sample Pearson correlation.

\subsection{Attribution-ensemble correction}\label{app:ensemble-correction}
Let $g(S,\zeta)=G(S)+\gamma(S,\zeta)$, with $\bE_\zeta[\gamma(S,\zeta)\mid S]=0$, and let the attribution members be i.i.d.\ and independent of the evaluation subsets and retrainings. Define the noise variance across subsets and its signal ratio by
\[
\sigma_\gamma^2\coloneqq
\bE_\zeta[\Var_{S\sim\piS}(\gamma(S,\zeta))],
\qquad
c\coloneqq\frac{\sigma_\gamma^2}{\Var_S(G)}.
\]
The variance across subsets removes member-specific offsets, which do not affect LDS. Independence of members gives the moment identities
\[
\bE_{\zeta_{1:E}}[\Var_S(\bar g_E)]
=\Var_S(G)+\frac{\sigma_\gamma^2}{E},
\qquad
\bE_{\zeta_{1:E}}[\Cov_S(\bar g_E,F)]
=\Cov_S(G,F).
\]
The ratio formed from these expected moments is therefore
\[
\widetilde\rho_E^{\,\mathrm p}
=\frac{\Corr_S(G,F)}{\sqrt{1+c/E}}.
\]
Comparing ensemble sizes $K$ and $E$ gives the main-text forecast. This approximates the correlation of a realised ensemble: increasing the number of subsets alone does not remove uncertainty from the shared ensemble members.

For $K\geq2$ observed members, paired differences cancel the shared signal and yield
\[
\widehat{\sigma}_\gamma^2
\coloneqq {K\choose2}^{-1}\sum_{i<j}
 \frac{\widehat{\Var}_M\!\left(g(S,\zeta_i)-g(S,\zeta_j)\right)}{2},
\qquad
\widehat{\sigma}_G^2
\coloneqq \widehat{\Var}_M(\bar g_K)-\widehat{\sigma}_\gamma^2/K.
\]
When $\widehat{\sigma}_G^2>0$, we use $\widehat c=\widehat{\sigma}_\gamma^2/\widehat{\sigma}_G^2$ in the forecast.

\subsection{Combining noise and finite-subset corrections}\label{app:combined-correction}
For \cref{fig:fig1a}, the ensemble experiments and \cref{fig:fig6-rmse,fig:fig6-single-draw}, let $\widehat s_U$ be the query average of \eqref{eq:sp_unb} applied to the observed prediction/readout pairs, and let $\widehat p_{\mathrm{corr}}$ and $\widehat p_{\mathrm{obs}}$ be the corresponding query-averaged corrected and ordinary Pearson correlations. With $h(p)=\frac6\pi\arcsin(p/2)$ and $\Pi$ denoting projection onto $[-1,1]$, the combined estimator is
\begin{equation}\label{eq:combined-correction}
\widehat\rho^{\,\mathrm{s,comb}}
=h\!\left(\Pi\!\left[
\frac{\widehat p_{\mathrm{corr}}}{\widehat p_{\mathrm{obs}}}
\,h^{-1}\!\left(\Pi[\widehat s_U]\right)
\right]\right).
\end{equation}
In the experiments, the corrected readout variance is floored at $\widehat{\Var}_M(\bar f_R)/4$. This safeguard is not part of Proposition~\ref{prop:corrected} and can retain attenuation when $\lambda/R>3$. The ratio factor is set to $1$ when $|\widehat p_{\mathrm{obs}}|\leq10^{-12}$. Averaging over queries precedes the transformations; the nonlinear combination is not guaranteed to be unbiased.

\section{Experimental Details}\label{app:experimental-details}

\subsection{Model training and retraining banks.}
We use precomputed retraining banks summarised below, which we release publicly alongside the code that produces them. Each correlation is computed per query before averaging over queries.

\begin{center}
\begin{tabular}{lrrr}
\toprule
Dataset / model & Subsets & Seeds per subset & Queries \\
\midrule
CIFAR-10 / ResNet-9 & 300 & 40 & 2,000 \\
QNLI / BERT & 400 & 20 & 2,000 \\
WikiText-2 / GPT-2 & 400 & 20 & 481 \\
\bottomrule
\end{tabular}
\end{center}

\paragraph{CIFAR-10/ResNet-9.}
We follow the training setup of \citet{Park2023TRAK}, using ResNet-9 and the full CIFAR-10 training set. Each retraining subset contains $25{,}000$ images ($\alpha=0.5$). Models are trained from scratch for $24$ epochs using SGD with momentum $0.9$, batch size $512$, and weight decay $5\times10^{-4}$. The learning rate increases linearly from zero to $0.5$ over the first five epochs, then decreases linearly to zero. Training uses random horizontal flips, translations, and cutout augmentation. Each subset--replica pair uses an independent training seed. For each query image, the readout is the correct-class log-odds margin.

\paragraph{QNLI/BERT.}
Following \citet{Park2023TRAK}, we restrict the QNLI training pool to $50{,}000$ examples and fine-tune pretrained \texttt{bert-base-cased} on subsets of $25{,}000$ examples ($\alpha=0.5$). We remove the pooler's $\tanh$ activation and fine-tune for $20$ epochs using SGD with batch size $8$. The learning rate starts at $10^{-3}$ and decreases linearly to zero without warmup. Each subset--replica pair uses an independent training seed. For each query example, the readout is the correct-class log-odds margin.

\paragraph{WikiText-2/GPT-2 small.}

We follow the setup of \citet{Bae2024SOURCE}, as implemented in Bergson \citep{Quirke2026Bergson}. Each retraining subset contains a random half of the $4{,}656$ training chunks ($\alpha=0.5$). For each subset and seed, we fine-tune GPT-2 small (124M parameters) from the base checkpoint using AdamW with a constant learning rate of $3\times10^{-5}$, batch size $8$, weight decay $0.01$, and three epochs in fp32. Dropout is enabled during training and disabled during evaluation. Each query is one of the $481$ validation documents, and its readout is the negative cross-entropy loss. On this bank, our EK-FAC (ten models) attains a five-seed Spearman LDS of $0.464$, close to the $0.468\pm0.015$ reported for this setup.

Following the Bergson implementation, the WikiText-2 experiment shares each retraining seed across all subsets, whereas CIFAR-10 and QNLI use independent seeds for each subset–replica pair. The pretrained GPT-2 checkpoint is fixed, while the seed controls data ordering and dropout. For a given query, the effect of a seed can include both an offset shared across subsets and a component that varies with the subset. The shared offset cancels in correlations across subsets. We therefore centre each seed’s readouts across subsets before estimating the seed-noise variance. Shared seeds can still induce dependence across subsets, whereas the bounds in \cref{app:complexity} assume i.i.d.\ subset observations.

\subsection{Attribution configurations}
\paragraph{TRAK.}
We use TRAK \citep{Park2023TRAK} with models trained on independently sampled halves of the training data. We featurize all model parameters, with projection dimensions $20{,}480$ for CIFAR-10, $4{,}096$ for QNLI, and $2{,}048$ for MNIST and WikiText-2. The model output used for featurization is the correct-class log-odds margin for classification and the sum of per-token log-odds for WikiText-2. For TRAK-$k$, each attribution member combines $k$ models by separately averaging the kernel scores and output-to-loss weights before taking their product. We then average the scores of $E$ such members, so the total number of models is $kE$. The ensemble experiments use $k\in\{1,10,20\}$ on CIFAR-10 and $k\in\{1,5\}$ on QNLI and WikiText-2.

\paragraph{EK-FAC.}
We compute influence function attributions using EK-FAC \citep{Grosse2023EKFAC}, implemented with the \texttt{kronfluence} package\footnote{\url{https://github.com/pomonam/kronfluence}}, at the final checkpoints of models trained on the full training set. In the case of TDA ensembles, each attribution member corresponds to one model, and we average their attribution scores. We use damping $10^{-8}$ and estimate the Fisher factors using labels sampled from each model's predictive distribution, with an independent factor seed for each member. The attribution banks contain $30$ members on CIFAR-10 and MNIST, $20$ on QNLI, and $10$ on WikiText-2. The computation includes the supported convolutional and linear layers; for GPT-2, these are the $48$ transformer linear layers, excluding embeddings and LayerNorms.

\paragraph{SOURCE.}
We use SOURCE \citep{Bae2024SOURCE} with $L=3$ segments and two checkpoints per segment on CIFAR-10 and MNIST. For CIFAR-10, the six checkpoints are saved every four epochs of a $24$-epoch full-data training run. For MNIST, we use checkpoints at epochs $3,6,10,13,16,20$. We compute one attribution estimate per training trajectory and average over up to $30$ trajectories. Thus, $L$ counts the segments within one member, and $E$ counts the trajectories being averaged. 

\paragraph{TracIn.}
We use TracInCP \citep{Pruthi2020TracIn}, combining checkpoint contributions within each full-data training trajectory and then averaging the resulting attribution scores over trajectories. The QNLI attribution bank contains $20$ trajectories with checkpoints at epochs $4,8,12,16,20$. We project gradients over all parameters to dimension $1{,}024$ and weight checkpoint contributions according to the learning-rate schedule over the corresponding intervals. On WikiText-2, we use the same ten training runs and $48$ transformer linear layers as EK-FAC, with six half-epoch checkpoints per run and a random $16{,}384$-dimensional projection at each checkpoint.

\subsection{Evaluation methodology}
This section describes the estimators and reference scores for noise-corrected LDS, finite-seed LDS, and ordinary Spearman LDS on a fixed bank. For the noise-correction and finite-seed reconstruction benchmarks, we partition the retraining seeds into disjoint reference and estimation sets. Reference scores use all training subsets and only the reference seeds; the evaluated estimators use only the estimation seeds. The subset bank, queries, and attribution ensemble are shared.

\paragraph{Noise-corrected LDS estimation.}
\Cref{fig:correction}(a,~d,~g) evaluates the error in estimating the noise-free Spearman LDS with a single attribution member ($E=1$, EK-FAC), comparing ordinary sample Spearman with the combined correction \eqref{eq:combined-correction}. The repetitions, subset grid, reference scores, and predicted plateaus follow the protocol of \cref{fig:fig1a} described below, with seed splits of $20$ reference and $20$ estimation seeds on CIFAR-10, and $5$ reference and $15$ estimation seeds on QNLI and WikiText-2. Naive estimates use $R\in\{5,10,20\}$ on CIFAR-10 and $R\in\{3,5,10\}$ on QNLI and WikiText-2; corrected estimates use $R\in\{2,5\}$.

In \cref{fig:fig1a,fig:fig6-rmse,fig:fig6-single-draw}, we evaluate the error in estimating the noise-corrected Spearman LDS. We compare ordinary sample Spearman with the combined correction \eqref{eq:combined-correction}. The reference applies the combined estimator to all subsets and the reference seeds, using the same fixed attribution ensemble as the evaluated estimates. This ensemble consists of all $30$ CIFAR-10 EK-FAC members in \cref{fig:fig1a}, and the first $10$ members of each method in \cref{fig:fig6-rmse,fig:fig6-single-draw}.

For each of $200$ repetitions, we split the seeds into equal reference and estimation sets: $20/20$ on CIFAR-10 and $10/10$ on QNLI and WikiText-2. At each budget $N=mR$, we sample $m$ of the bank's subsets and $R$ seeds from the estimation set, uniformly without replacement. The base grid is $m\in\{5,8,12,25,50,100,200,300\}$, capped at the bank size, with additional points at $m=2$ for naive estimates and $m=3$ for corrected estimates, retaining budgets $mR\geq9$. Naive estimates use $R=5,10,20$ on CIFAR-10 and $R=5,10$ on QNLI and WikiText-2; corrected estimates use $R=2,5$ throughout. In the appendix experiments, all attribution methods within a dataset use the same splits and draws. \Cref{fig:fig6-rmse} reports RMSE against each repetition's reference, and \cref{fig:fig6-single-draw} shows the first repetition with its corresponding reference.

In \cref{fig:fig1a}, each curve shows the mean LDS estimate over the $200$ repetitions, and the black line shows the mean reference across seed splits. The shaded bands are centred on the mean estimate and extend by one sample standard deviation of the signed errors relative to each repetition's reference. The additional naive $R=40$ curve uses all $40$ CIFAR-10 retraining seeds, with evaluation subsets sampled without replacement. For this curve, errors are measured against the fixed combined-estimator reference computed from all $300$ subsets and all $40$ seeds, so its reference shares seeds with the estimates. Its band therefore measures only variability from subset sampling.

Writing $\widehat\rho_{\mathrm{ref}}^{\mathrm s}$ for a split's reference score, $\widehat\lambda$ for the median per-query noise-to-signal ratio estimated from its reference seeds, and $h(r)=(6/\pi)\arcsin(r/2)$, the predicted finite-seed plateau is
\[
h\!\left(\frac{h^{-1}(\widehat\rho_{\mathrm{ref}}^{\mathrm s})}{\sqrt{1+\widehat\lambda/R}}\right).
\]
The predicted RMSE floor is the absolute difference between this plateau and the reference score. We average these floors over seed splits for the RMSE plots; the single-draw plot uses its own split's reference and plateaus. Variability in the reference estimates contributes to the reported RMSE.

\paragraph{Finite-seed LDS reconstruction.}
In \cref{fig:correction}(b,~e,~h), the target is the standard five-seed Spearman LDS, which we estimate using either naive five-seed averaging or reconstruction from $k=2$ seeds. These experiments use a single attribution member ($E=1$). The reconstructed estimate applies \eqref{eq:finite-seed-reconstruction} to each query, converts the result to Spearman using \eqref{eq:pearson-spearman}, and averages over queries.

We use $30$ random seed splits into equal reference and estimation sets ($20/20$ on CIFAR-10 and $10/10$ on QNLI and WikiText-2), cycling through these splits over $300$ repetitions. For each split, the reference is evaluated on all subsets and averages Spearman LDS over $40$ random groups of five reference seeds. Each repetition samples subsets and estimation seeds without replacement and is scored against its split's reference. The subset counts are $m\in\{5,10,20,40,60,100\}$ for all three datasets. The budgets are $N=5m$ for naive averaging and $N=2m$ for reconstruction. We compute the reported budget ratios from log--log fits at RMSE $0.01$, except for QNLI/TracIn, where we use $0.0134$, the lowest RMSE reached by both curves.

\paragraph{Ensemble forecasting.}
In \cref{fig:fig1b} and \cref{fig:correction}(c,~f,~i), we compare forecasts based on a small attribution ensemble with measured noise-corrected LDS as the ensemble size increases. We use all subsets in each retraining bank and split the retraining seeds into equal halves, anchoring the forecast on one half and measuring on the other. The forecast uses the last $K$ attribution members. We map their combined Spearman estimate \eqref{eq:combined-correction} to the Pearson scale using $h^{-1}$, apply \eqref{eq:finite-ensemble-prediction}, and convert the forecast back using $h$. We estimate the attribution noise-to-signal ratio from differences between members as described in \cref{app:ensemble-correction}, and take the median of the finite per-query estimates of $c$.

\paragraph{Finite-subset debiasing.}
In \cref{fig:fig1c}, we fix the predictions from $100$ TRAK-1 members and average the readouts over all $40$ retraining seeds. We then sample subsets from the $300$-subset CIFAR-10 bank and compare ordinary Spearman with the finite-bank correction \eqref{eq:finite-bank-spearman}. The reference is ordinary Spearman computed on all $300$ subsets, averaged over the $2{,}000$ queries. We use $m\in\{4,5,6,8,12,16,24,32,48,64,96,128\}$, with $2{,}500$ uniform draws without replacement at each subset count. Both estimators use the same draws. Relative RMSE is obtained by dividing RMSE by the absolute reference LDS.

\Cref{fig:bias_variance} decomposes the same squared error into squared bias and variance. In \cref{fig:bias_variance_queries}, we use the same subset draws and vary the number of queries, taking $Q=1,100,1000$ from nested prefixes of one random permutation of the queries. Each reference score is computed over the corresponding query set.

\Cref{fig:spearman_grid} extends the comparison to TRAK-1 with $E=100$, EK-FAC with $E=30$, and SOURCE with $L=3$ and $E=30$, on CIFAR-10 and MNIST. The MNIST bank contains $300$ subsets, $40$ retraining seeds per subset, and $2{,}000$ queries. It uses a training pool of $6{,}144$ examples, half-sized retraining subsets, and an MLP with three hidden layers of widths $512,256,128$. The models are trained for $20$ epochs using SGD with momentum $0.9$, learning rate $0.03$, batch size $64$, and weight decay $0.001$. For both datasets, we fix the first five retraining seeds and use $600$ subset draws at each value of $m$ in the same grid as \cref{fig:fig1c}. We compare ordinary Spearman with \eqref{eq:finite-bank-spearman}, using ordinary Spearman on all $300$ subsets with these fixed five seeds as the reference.

\section{Analysis of the Pearson-Spearman conversion}\label{sec:normality}

We assess the accuracy of the Pearson--Spearman conversion \eqref{eq:pearson-spearman} in a separate reconstruction diagnostic on CIFAR-10, QNLI and WikiText-2, with $E=1$. For the TRAK results, we use attribution scores computed from a single model (TRAK-1, with $E=1$). Seven of the nine estimated reconstruction biases are smaller in magnitude than the standard deviation across two-seed reconstructions.

\begin{figure}[H]
  \centering
  \includegraphics[width=0.95\linewidth]{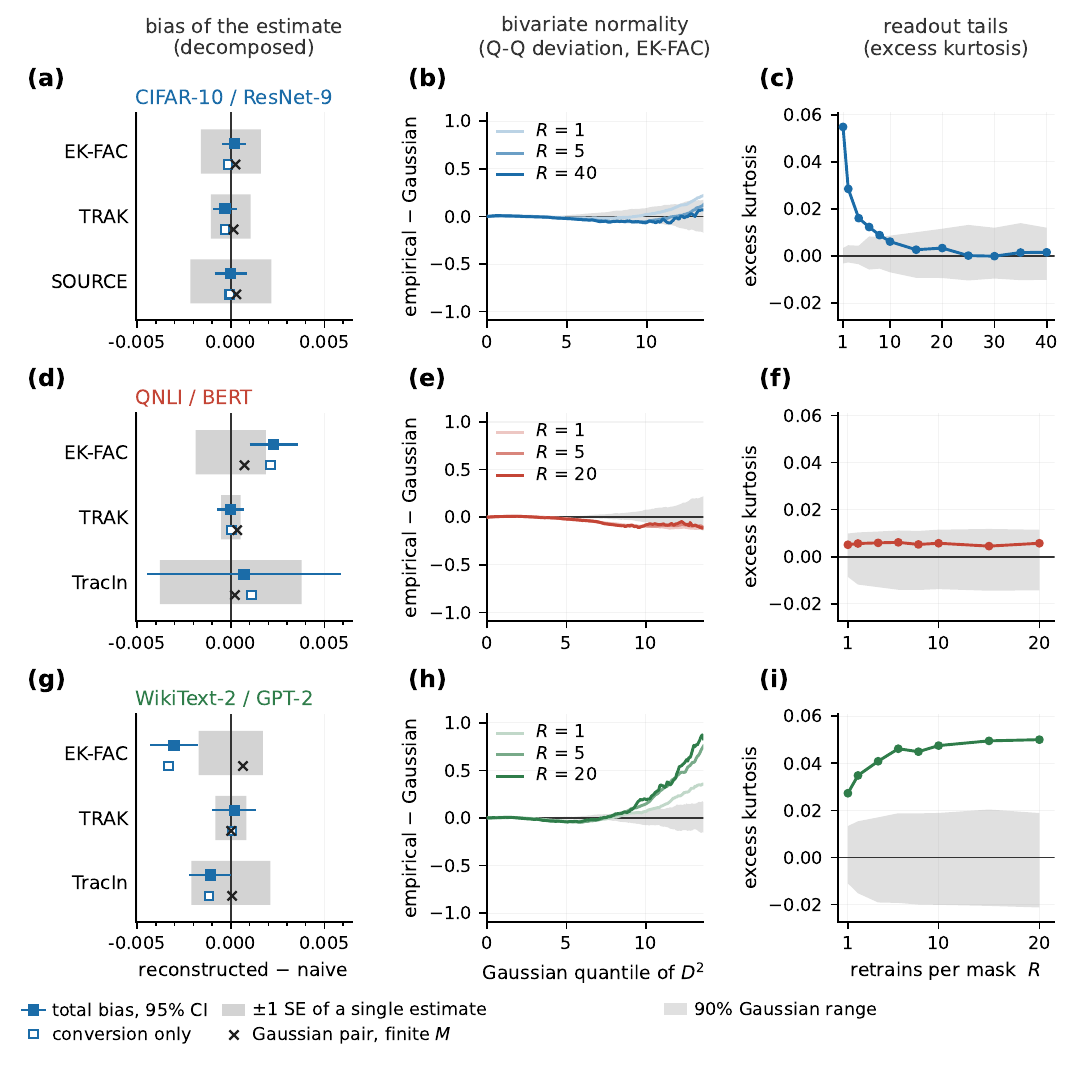}
  \captionsetup{font=small}
  \caption{\textbf{The arcsin bridge in detail.}
  Top: CIFAR-10 / ResNet-9; middle: QNLI / BERT; bottom: WikiText-2 / GPT-2; single
  attribution members ($E = 1$).
  \textbf{(a, d, g)}~Bias of the reconstructed estimate (filled squares, 95\% CI), its
  conversion component (open squares), and the corresponding finite-mask Gaussian
  baseline (crosses); grey: $\pm 1$ standard deviation across two-seed reconstructions.
  \textbf{(b, e, h)}~Q--Q deviation of $D^2$ for EK-FAC, with the readout averaged over $R$
  retrains per mask; 0 denotes the Gaussian reference.
  \textbf{(c, f, i)}~Excess kurtosis of the readout relative to the mean of its
  rotation-based reference distribution, against $R$.
  Grey in (b, c, e, f, h, i): pointwise 90\% rotation-based reference ranges.}
  \label{fig:qq-normality}
\end{figure}

\paragraph{Reconstruction and conversion error (a, d, g).}
We estimate reconstruction bias as the difference between the two-seed reconstruction and the naive five-seed Spearman score, evaluated on the full subset bank. To isolate conversion error, we compare the transformed five-seed sample Pearson correlation with sample Spearman, averaging over queries. The Gaussian arcsine relation holds for population correlations. With only $M$ sampled subsets, the transformed sample Pearson correlation and sample Spearman correlation can differ, including in expectation, even when the attribution--readout pairs are jointly Gaussian. The largest observed magnitude of the additional discrepancy relative to this Gaussian baseline is $0.004$ (WikiText-2 EK-FAC), and at most $0.0014$ on CIFAR-10 and QNLI.

\paragraph{Distributional diagnostics (b, c, e, f, h, i).}
On CIFAR-10, the heavier upper tail in the Q--Q diagnostic and the excess readout kurtosis diminish as retraining seeds are averaged. QNLI retains Q--Q departures, while its readout kurtosis remains within the Gaussian reference range. On WikiText-2, both the Q--Q departure and the readout excess kurtosis grow as seeds are averaged and stay outside the Gaussian reference ranges. The distance diagnostic assesses only the radial distribution; neither diagnostic establishes full bivariate normality or Gaussianity of the noise-free pair.

\paragraph{Seed averaging.}
We estimate the reconstruction discrepancy on the full bank of evaluation subsets. We average the reconstructed five-seed Spearman score over $200$ randomly drawn groups of two retraining seeds, and the naive five-seed Spearman score over $40$ randomly drawn groups of five retraining seeds. Each score is averaged over queries. Seeds within each group are sampled without replacement from the full available seed pool, with no separate reference seeds reserved.

\paragraph{Uncertainty estimates.}
We quantify uncertainty in the estimated reconstruction discrepancy using a delete-a-group jackknife over evaluation subsets. We randomly partition the subsets into $20$ groups and recompute both averaged scores after omitting each group. The resulting jackknife standard error gives an approximate 95\% confidence interval, formed by adding and subtracting $1.96$ standard errors from the estimated discrepancy. Separately, we report the standard deviation across the $200$ two-seed reconstructions to show the variability of a single reconstruction on the fixed subset bank.

\paragraph{Gaussian reference diagnostics.}
For each query, we compute the squared Mahalanobis distance $D^2$ of each attribution--readout pair $(G(S_m),\bar f_R(S_m))$ from the sample mean, using the sample covariance with divisor $M-1$. Under independent bivariate-normal sampling,
\[
\frac{M D^2}{(M-1)^2}\sim\mathrm{Beta}\!\left(1,\frac{M-3}{2}\right).
\]
We pool these distances over queries and plot their empirical quantiles minus the Gaussian reference quantiles for $R=1$, $5$, and all available seeds. The shaded Q--Q envelope gives the pointwise 5th and 95th percentiles of $40$ rotation-based reference curves at $R=5$. To generate each reference curve, we centre the attribution predictions and seed readouts across subsets, apply the same random orthogonal rotation across the $M$ subset observations to all queries and readouts, and recompute the sample means, covariances and distances. For readout excess kurtosis, we average over queries and subtract the mean of the corresponding rotation-based reference distribution at each $R$; the shaded range gives its 5th and 95th percentiles after the same subtraction.

\section{Additional Results for Noise Correction}\label{sec:additional-results}
\begin{figure}[H]
  \centering
  \includegraphics[width=\linewidth]{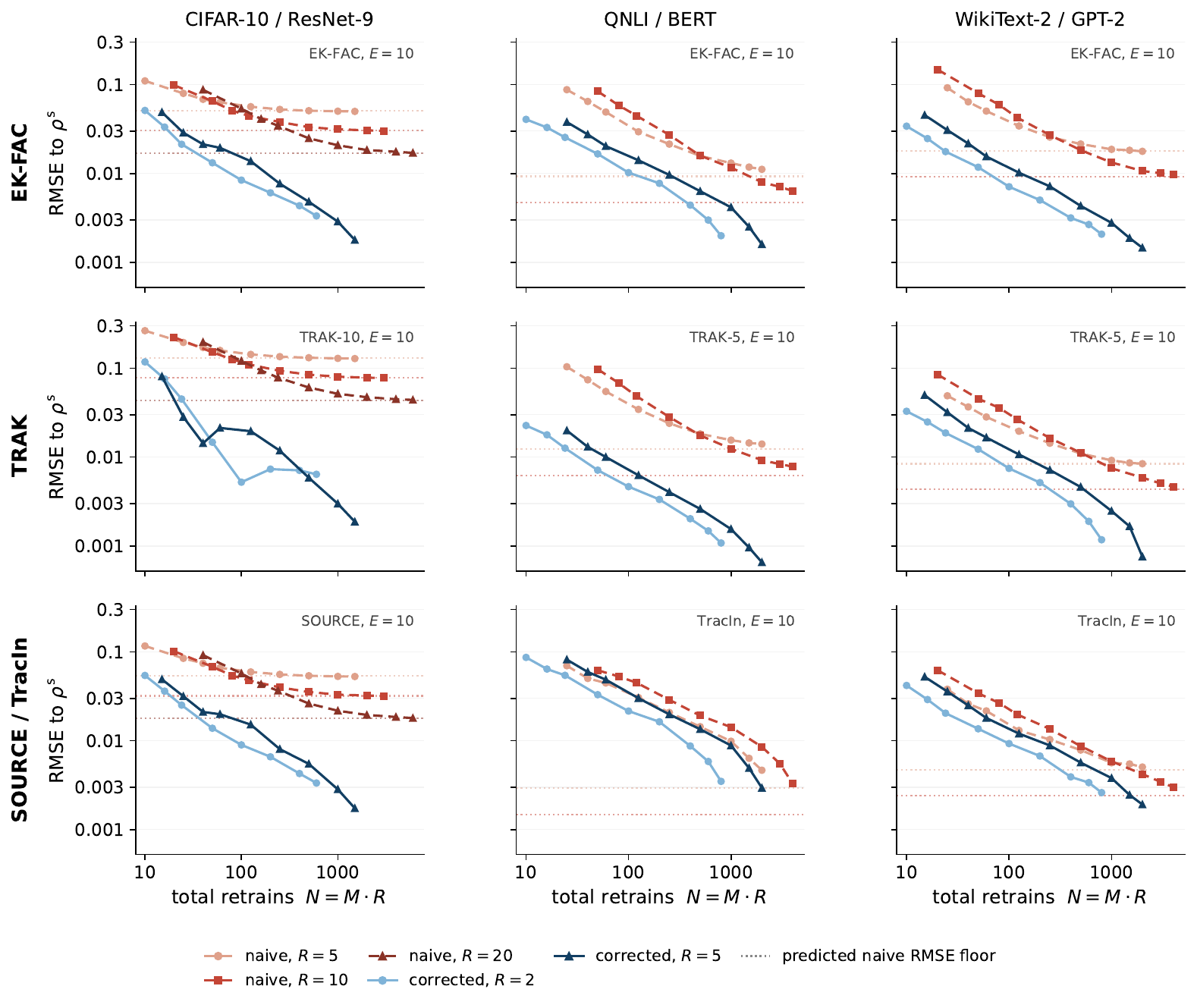}
  \captionsetup{font=small}
  \caption{\textbf{RMSE of corrected Spearman LDS.}
  Error to references computed from held-out retraining seeds over $200$ draws, with $E=10$.
  Columns: CIFAR-10, QNLI, WikiText-2. Rows: EK-FAC, TRAK, SOURCE/TracIn.
  Dashed: naive Spearman; solid: combined training-noise and finite-subset correction;
  dotted: predicted naive RMSE floors.}
  \label{fig:fig6-rmse}
\end{figure}

\begin{figure}[H]
  \centering
  \includegraphics[width=\linewidth]{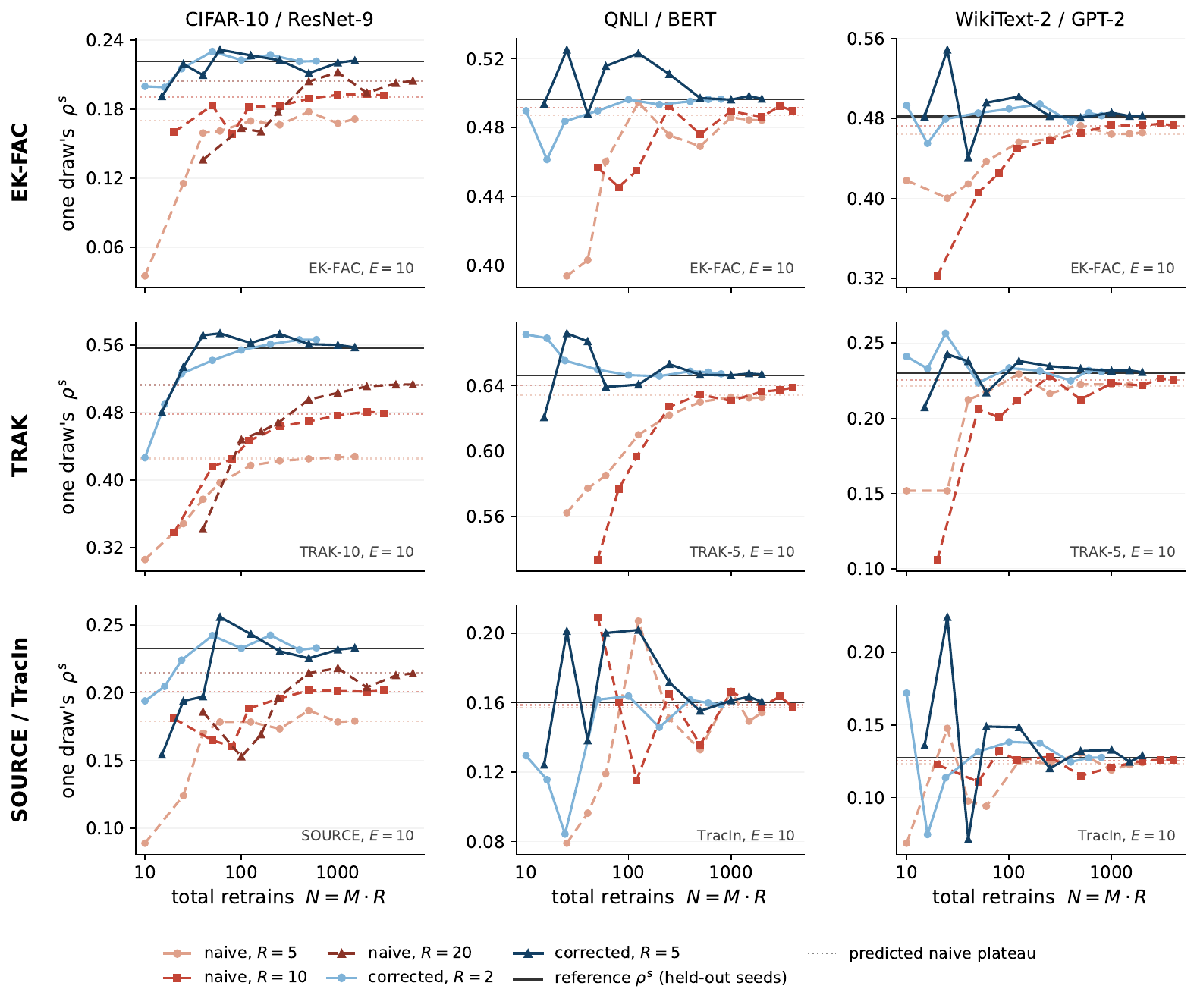}
  \captionsetup{font=small}
  \caption{\textbf{Spearman LDS estimates from a single draw.}
  The first of the $200$ draws in \cref{fig:fig6-rmse}, with the same $E=10$
  ensembles and estimators. Black lines: reference computed from the first split's held-out retraining seeds;
  dotted lines: predicted naive plateaus from that split.}
  \label{fig:fig6-single-draw}
\end{figure}

\begin{figure}[H]
    \centering
    \includegraphics[width=\linewidth]{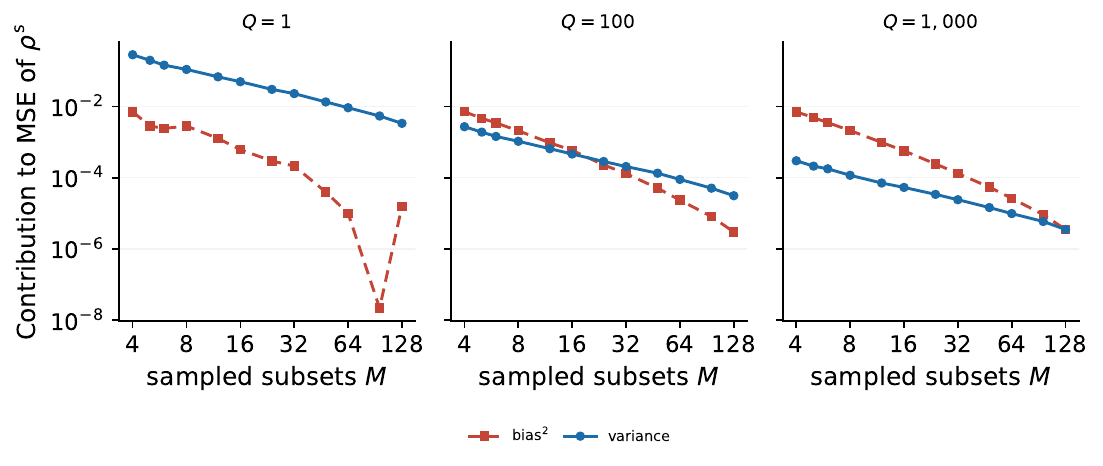}
    \caption{Comparing the bias (squared) and variance of estimating $\rho^s$ using the finite-sample Spearman correlation estimator $\empsp$, across different numbers of query points $Q$.}
    \label{fig:bias_variance_queries}
\end{figure}

\begin{figure}[H]
    \centering
    \includegraphics[width=\linewidth]{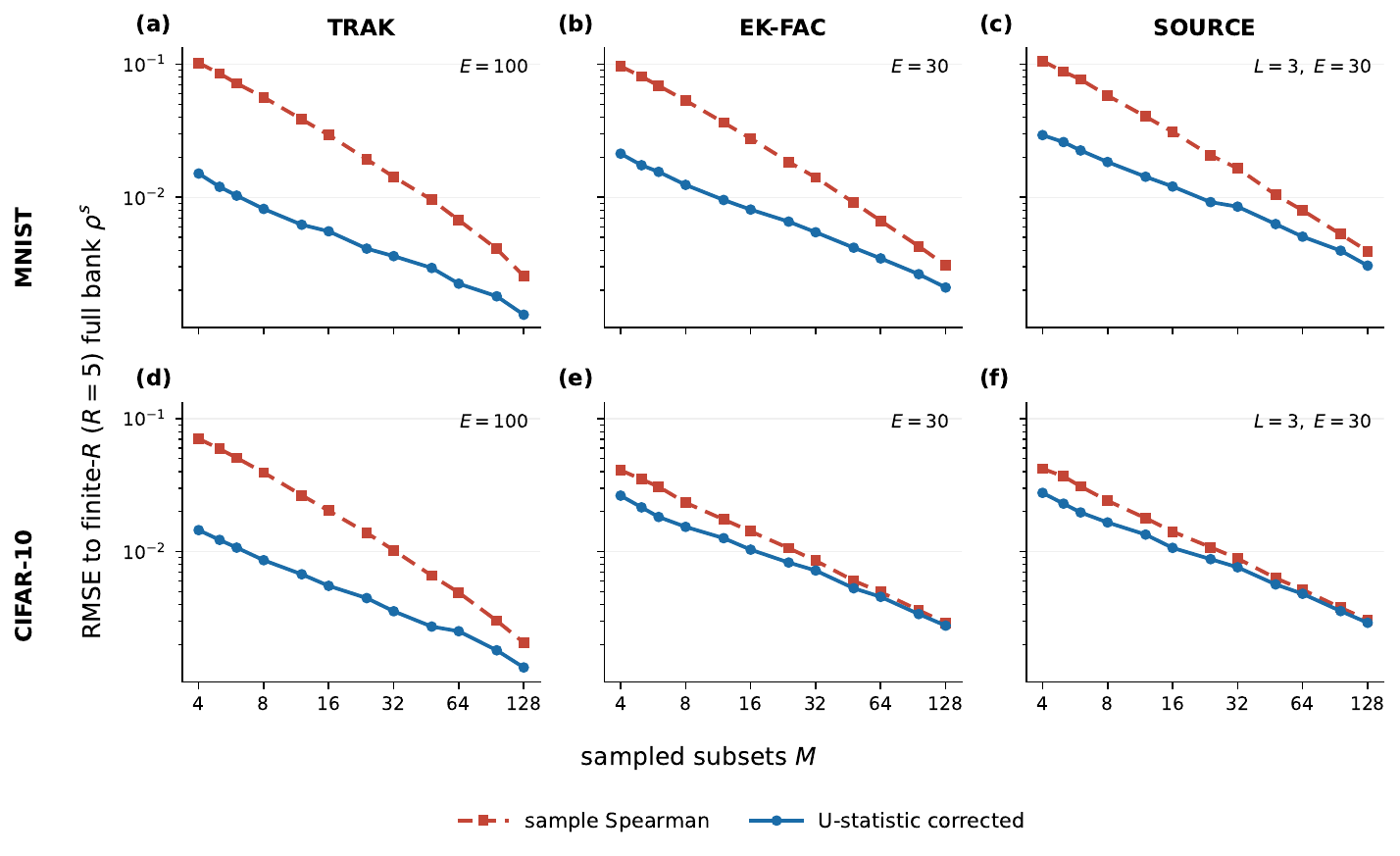}
    \caption{Comparing RMSE in estimating the full-bank LDS using finite-sample Spearman and the finite-bank correction \eqref{eq:finite-bank-spearman}. All experiments use $R=5$ retraining seeds.}
    \label{fig:spearman_grid}
\end{figure}

\Cref{fig:fig6-rmse,fig:fig6-single-draw} show the effect of combining training-noise and finite-subset corrections on CIFAR-10, QNLI and WikiText-2 with $E=10$ attribution members. Here, we compare estimates against a noise-corrected Spearman reference computed from held-out retraining seeds, whereas \cref{fig:correction}(b,~e,~h) targets the biased five-seed score. Correction with $R=2$ or $R=5$ reduces RMSE as the retraining budget grows. The naive attenuation floors are largest on CIFAR-10; on QNLI and WikiText-2, the smaller floors make sampling variability more prominent. The reference and sampling procedure are described in \cref{app:experimental-details}.

\clearpage
\subsection{Additional Results for Finite-Sample Debiasing}
This section contains additional figures showing the bias and variance in estimating the LDS, and the effect of the unbiased U-statistic estimator.

\end{document}